\pdfoutput=1

\documentclass[11pt]{article}

\usepackage[]{EMNLP2023}
\usepackage{booktabs} 
\usepackage{times}
\usepackage{latexsym}
\usepackage{multirow}
\usepackage{subcaption}

\usepackage[T1]{fontenc}

\usepackage[utf8]{inputenc}

\usepackage{microtype}

\usepackage{inconsolata}
\usepackage{amsmath}

\usepackage{graphicx}

\title{
TopicAdapt- An Inter-Corpora Topics Adaptation Approach
}

\author{Pritom Saha Akash $\quad$ Trisha Das $\quad$ Kevin Chen-Chuan Chang\\
University of Illinois at Urbana-Champaign, USA \\
 \texttt{\{pakash2, trishad2, kcchang\}@illinois.edu}
}

\begin{document}
\maketitle

\begin{abstract}
Topic models are popular statistical tools for detecting latent semantic topics in a text corpus. They have been utilized in various applications across different fields. However, traditional topic models have some limitations, including insensitivity to user guidance, sensitivity to the amount and quality of data, and the inability to adapt learned topics from one corpus to another. To address these challenges, this paper proposes a neural topic model, TopicAdapt, that can adapt relevant topics from a related source corpus and also discover new topics in a target corpus that are absent in the source corpus. The proposed model offers a promising approach to improve topic modeling performance in practical scenarios. Experiments over multiple datasets from diverse domains show the superiority of the proposed model against the state-of-the-art topic models.
\end{abstract}
\section{Introduction}
\label{sec:introduction}
To effectively and quickly comprehend and navigate a big text corpus, it is important to mine a set of diverse and cohesive topics automatically. Topic models \cite{jordan1999introduction,blei2003latent} are statistical tools for detecting latent semantic themes in a text collection. These approaches have gained popularity for text mining \cite{foster2007mixture,mei2007automatic} and information retrieval tasks \cite{dou2007large,wei2006lda} spanning a wide range of applications in fields such as science, humanities, business, and other related areas \cite{boyd2017applications}. 

Despite the effectiveness of standard topic models for understanding latent topics in a large corpus, they suffer from several drawbacks. 
Firstly, the traditional topic \cite{jordan1999introduction, blei2003latent} models do not consider user guidance in learning the topics. For example, users may already know the name of topics but want to know the corpus-specific representation of that topic. 

Secondly, the performance of topic models is often sensitive to the amount of data and the quality of the data, and a small corpus may not provide enough information to identify the underlying topics accurately. One possible way to handle this is to adapt a pre-trained topic model from a related corpus to the target corpus. It leverages the knowledge learned from a large source corpus to improve the topic modeling performance on the small target corpus. However, in traditional topic models, there is no specific way to adapt learned topics from one corpus to another.

Moreover, not all the topics of the related source domain are actual topics of the target domain, and there may also exist new topics in the target corpus different from the source corpus. For example, the source domain may cover topics such as ``politics'' and ``sports'' where the target may have a new topic, ``entertainment'', different from the source domain. Therefore, we develop a model named \textbf{TopicAdapt} that can dynamically adapt relevant topics from the source domain by transfer learning and also can discover new topics available in the target domain but absent in the source domain. To evaluate the performance of the proposed model, we conduct both quantitive and qualitative evaluations over multiple datasets from diverse domains. The experimental results show the superiority of the proposed model against the state-of-the-art topic models.

\section{Methodology}
\label{sec:method}
\subsection{Problem Statement}

We proposed a problem of adapting topics from one corpus (i.e., domain) to another. As input, it takes a target corpus $\mathcal{D}$, the topic-word distribution from a source reference corpus or alternatively named representation $\beta^r$ for $k$ well-defined topics with their surface names $\mathcal{C}$. As output, we want to learn topic-word distribution $\beta$ for the target corpus that best represents the corpus by given well-known topics. It also aims to generate new topics from the target corpus without any supervision or minimal supervision, such as using only the topic surface names. Similar to an existing topic model named Coordinated Topic Model (CTM) \cite{akash2022coordinated}, we can get a set of well-defined topics with their representation. More specifically, we use labeled LDA  \cite{ramage2009labeled} to get reference representation (more details on \cite{akash2022coordinated}).

\begin{figure}
\centering
\includegraphics[width=1.0\columnwidth]{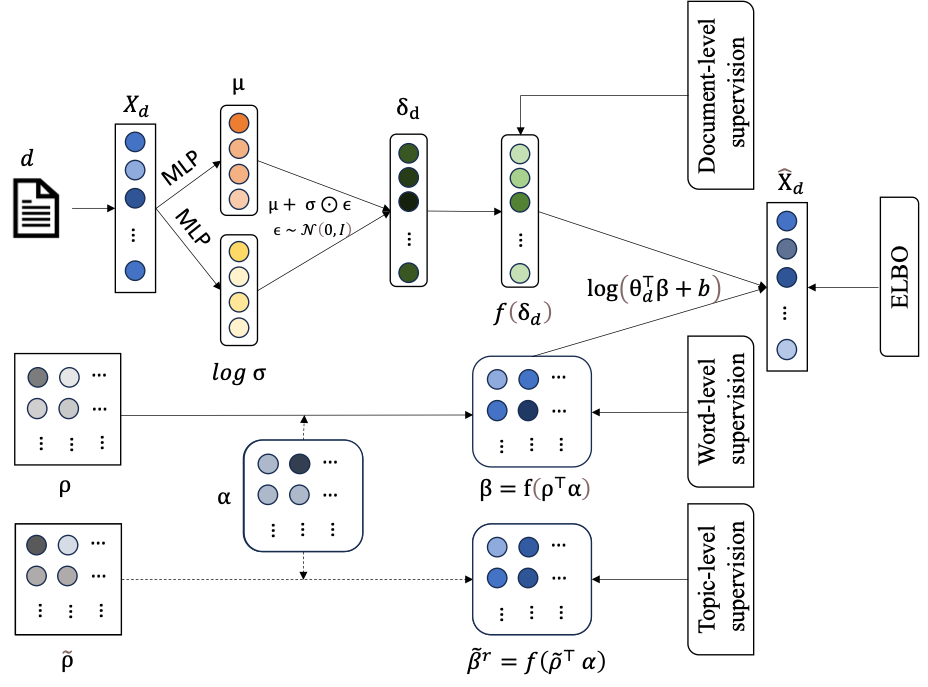}
\caption{Proposed Architecture}
\label{fig:arch}
\vspace{-6mm}
\end{figure}

To solve our problem, we have chosen to use the Embedded Topic Model (ETM) \cite{dieng2020topic} as the foundation of our proposed model and extension of a recent topic model named Coordinated Topic Modeling (CTM) \cite{akash2022coordinated}. We have several compelling reasons for this choice. Firstly, ETM is an excellent choice because it effectively combines the strengths of neural topic modeling and word embedding when modeling a corpus. Secondly, using pre-trained word embeddings enables us to map words in a common vector space, even if those words are not present in the target corpus vocabulary. Finally, we can impose our problem-specific requirements by applying regularization techniques to the objectives of ETM. Similar to CTM, the proposed framework uses topic-level and document-level supervision. Moreover, our model also incorporates word-level supervision for having topics comprising semantically similar words.  

As part of our problem, we are given a topic-word distribution $\beta^r$ for some known topics, along with their surface names $\mathcal{C}$. We aim to adapt these topics for a target corpus $\mathcal{D}$ and discover new topics. To achieve this, we have modified the ETM model to incorporate supervision from $\mathcal{C}$ and $\beta^r$ as guidance. However, we cannot directly use $\beta^r$ in the ETM model for the target corpus due to the vocabulary mismatch problem with the reference corpus. Therefore, we have modified the original ETM model structure, similar to CTM, to learn a topic-word distribution with vocabulary dimensions comparable to $\beta^r$. Additionally, we have generated pseudo-labeled documents in the target corpus using $\mathcal{C}$ to enhance document modeling in ETM. Lastly, we have also used $\mathcal{C}$ to bias the topic distribution and create topics consisting of semantically closer words. The overview of our model is shown in Figure \ref{fig:arch}.

\subsection{TopicAdapt}
\textbf{Topic-level Supervision:}
A set of topics with a reference representation $\beta^{r}$ is employed as source topics to guide the generation of a target representation $\beta$ that best captures the characteristics of the given $\mathcal{D}$. The reference representation may be obtained from sources such as a large annotated corpus in a similar domain. However, a key issue arises in using $\beta^{r}$ directly as guidance, as it cannot be assumed that $\beta^{r}$ and $\beta$ share the same vocabulary. To solve this, following CTM, \cite{akash2022coordinated}, an indirect method of supervision called "reference projection" is employed. To elaborate further, in conjunction with the parameter $\beta$, the projected representation $\tilde{\beta}^{r} = f(\tilde{\rho}^\top \alpha)$ is computed where $\tilde{\rho}$ denotes the embedding matrix associated with the lexicon upon which the reference $\beta^{r}$ is constructed. Finally, $\tilde{\beta}^{r}$ is used to indirectly guide $\beta$ by minimizing the following:
\vspace{-4mm}
\begin{align*}
    R_\beta & = \frac{1}{k} \sum^k_{j= 1} KL(\beta^{r}_j,\tilde{\beta}^{r}_j)
\vspace{-5mm}
\end{align*}

\noindent \textbf{Document-level Supervision:}
 Similar to CTM \cite{akash2022coordinated}, in this study, we utilize $\mathcal{C}$ to obtain $\theta^{t}$ for document-level supervision. To achieve this, we employ a pre-trained textual entailment model \cite{liu2019roberta}. The model takes an input document $d$ as the ``premise'' creates a ``hypothesis'' by filling a template with a surface name $c_k \in \mathcal{C}$, and produces a probability $p_{dk}$ representing the extent to which the premise entails the hypothesis. This distribution is then used to guide document topic distribution. We directly utilize the generated probabilities $p_{dk}$ as a soft label for $\theta^t_{dk}$. Soft labeling offers the opportunity to implement a technique proposed by \cite{bhatia2016automatic}, which emphasizes the high-probability label while diminishing the low-probability ones. To accomplish this, the method squares and normalizes the $p_{dk}$ values in the following manner:
 \vspace{-3mm}
\begin{align*}
    \theta^t_{dk} &= \frac{p^2_{dk}/f_k}{\sum_{k'}p^2_{dk'}/f_{k'}}, f_k = \sum_{d \in D} p_{dk}
\end{align*}

The $\theta^{t}$ value is employed to offer supervision at the document level by reducing the following:
 \vspace{-2mm}
\begin{align*}
    R_\theta &= \frac{1}{|D|} \sum_{d \in D} KL(\theta^t_{d}, \theta_{d}).
\end{align*}

\noindent \textbf{Word-level Supervision:}
The distribution of topics over vocabulary words is such that the most relevant words in a given topic are semantically related to the topic's name. To leverage this observation, pretrained word embeddings are employed to obtain embeddings for all vocabulary words. Subsequently, cosine similarity between the surface name of a topic and vocabulary words is used to generate a topic conditional probability distribution over all the vocabulary words $\gamma$. $\gamma$ serves as a guide for constructing the topic-word distribution.
\vspace{-3mm}
\begin{align*}
    R_{\gamma} & =  \frac{1}{k} \sum^k_{j= 1} KL(\gamma_j,\beta_j)
\end{align*}

\subsubsection{Training}
We unify topic-level, document-level, and word-level supervision into one model by constraining the objective of our base model as follows:
\begin{align} \label{eq5}
\mathcal{L}(\theta) &= ELBO - \gamma_{\beta}R_{\beta}- \gamma_{\theta}R_{\theta} -\gamma_{\gamma}R_{\gamma},
\end{align}
where $\gamma_{\beta}$, $\gamma_{\theta}$ and $\gamma_{\gamma}$ are the regularization weights
for $R_{\beta}$, $R_{\theta}$ and $R_{\gamma}$ respectively. Maximizing Eq. \ref{eq5} ensures the following objectives: (1) The ELBO part enforces the model to explain $D$ by reducing the reconstruction error; (2) $R_{\beta}$ enforces the model to move $\beta$ in the direction of $\beta^r$; (3) $R_{\theta}$ encourages the model to maintain the global semantics of given topics in $\beta$ by enforcing $\theta$ and $\theta^{t}$ as similar as possible; and (4) $R_{\gamma}$ enforces topic words to be similar to relevant words in the vocabulary.
\section{Experiments}
\label{sec:experiments}
\subsection{Data}
We use three datasets from news articles: 20 Newsgroup corpus \footnote{http://qwone.com/~jason/20Newsgroups/}, New York Times annotated corpus \cite{sandhaus2008new}, AG’s News dataset \cite{yang2016hierarchical}. For the review sentiment domain, we use the Yelp restaurant review dataset and IMDB Movie Review dataset. For academic articles, we use: Arxiv abstracts \footnote{https://www.kaggle.com/Cornell-University/arxiv}, Microsoft Academic Graph AI article abstracts \cite{sinha2015overview}. See Appendix \ref{sec:data_detail} for more details.

\subsection{Baselines}
We compare our model with the following
baselines. GLDA \cite{jagarlamudi2012incorporating}, Sup+LLDA \cite{ramage2009labeled}, ZS+LLDA \cite{ramage2009labeled},  ACorEx, AVIAD \cite{hoang2019towards}, KeyETM \cite{harandizadeh2022keyword}, ECTM \cite{saha2022coordinated}. The details of the baselines can be found in Appendix \ref{sec:baseline_detail}.


\subsection{Topic Quality Evaluation}
We use the following three quantitative measurements to evaluate the quality of inferred topics: Topic coherence (TC), Topic diversity (TD), and Topic Quality (TQ). Details about these metrics can be found in the Appendix \ref{sec:eval_metric}.

We first show the quantitative
results of topic quality in Table \ref{tab:topic_quality}. The results suggest that, for news and sentiment domains, TopicAdapt generates more coherent and interpretable topics than other baselines. 

\begin{table}[!tb]
\centering
\resizebox{\linewidth}{!}{%
\setlength\tabcolsep{2pt}
\begin{tabular}{@{}c|ccc|ccc|ccc|ccc@{}}
\toprule
\multirow{2}{*}{Methods} & \multicolumn{3}{c|}{\textbf{20Newsg}}     & \multicolumn{3}{c|}{\textbf{NYT}}         & \multicolumn{3}{c|}{\textbf{Yelp-Senti}}  & \multicolumn{3}{c}{\textbf{Arxiv-AI}}    \\  
                         & TC & TD & TQ & TC & TD & TQ & TC & TD & TQ & TC & TD & TQ \\ \midrule
                         
GLDA                     & 0.25      & 0.87      & 0.22    & 0.26      & 0.85      & 0.22    & 0.08      & 0.80      & 0.06    & 0.09      & 0.93      & 0.09    \\
Sup+LLDA                    & 0.23      & 0.79      & 0.18    & 0.20      & 0.63      & 0.12    & 0.06      & 0.70      & 0.04    & 0.04      & 0.46      & 0.02    \\

ZS+LLDA                    & 0.23      & 0.80      & 0.18    & 0.17      & 0.65      & 0.11    & 0.06      & 0.76      & 0.05    & 0.14      & 0.80      & 0.11    \\

ACorEX                   & 0.25      & \textbf{1.00}       & 0.25    & 0.27      & \textbf{1.00}       & 0.27    & 0.07      & \textbf{1.00}       & 0.07    & -0.03      & 0.96       & -0.03    \\
AVIAD                    & 0.13      & \textbf{1.00}       & 0.13    & -0.26     & \textbf{1.00}       & -0.26   & -0.01      & \textbf{1.00}       & -0.01    & -0.34     & \textbf{1.00}       & -0.34   \\
KeyETM                   & 0.26      & \textbf{1.00}       & 0.26    & 0.19      & 0.89      & 0.17    & 0.07      & 0.92      & 0.07    & 0.04      & \textbf{1.00}       & 0.04    \\ \midrule
ECTM                     & \textbf{0.30}      & \textbf{1.00}       & \textbf{0.30}    & \textbf{0.28}      & 0.97      & \textbf{0.27}    & \textbf{0.09}      & \textbf{1.00}       & \textbf{0.09}    & \textbf{0.15}      & 0.97       & \textbf{0.15}    \\ 
\midrule

TopicAdapt & \textbf{0.30}      & \textbf{1.00}       & \textbf{0.30}    & \textbf{0.28}      & 0.97      & \textbf{0.27}    & \textbf{0.11}      & \textbf{1.00}       & \textbf{0.11}    & 0.12      & 0.93       & 0.12\\

\bottomrule
\end{tabular}}
\vspace{-2.0mm}
\caption{Quality Measures of Topic}
\label{tab:topic_quality}
\vspace{-5.00mm}
\end{table}
In Table \ref{tab:qualitative_short}, we show randomly selected two topics
from each dataset and top-5 words under each topic from reference topic words, ECTM and TopicAdapt. Words that we found to be irrelevant to the corresponding topic are marked
with ($\times$) in Table \ref{tab:qualitative_short}. The table consisting of results from all baselines can be found in \ref{sec:qual_eval}.

\begin{table*}[]
\centering
\resizebox{0.8\linewidth}{!}{%
\begin{tabular}{@{}c|cc|cc|cc|cc@{}}
\toprule
\multirow{2}{*}{} & \multicolumn{2}{c|}{\textbf{20Newsg}}                                                                                                                                                                    & \multicolumn{2}{c|}{\textbf{NYT}}                                                                                                                                                                                     & \multicolumn{2}{c|}{\textbf{Yelp-Senti}}                                                                                                                                                                                  & \multicolumn{2}{c}{\textbf{Arxiv-AI}}                                                                                                                                                                                            \\ 
                         & sports                                                                                         & politics                                                                                      & business                                                                                       & technology                                                                                                 & good                                                                                                    & bad                                                                                                   & ML                                                                                                                   & IR                                                                                               \\ \toprule
\begin{tabular}[c]{@{}c@{}} Reference\\Topic\\ Words\end{tabular}                    & \begin{tabular}[c]{@{}c@{}}night\\play\\sport\\player\\beat\end{tabular}                 & \begin{tabular}[c]{@{}c@{}}leader\\ election \\ attack\\ afp\\ iraqi \end{tabular}       & \begin{tabular}[c]{@{}c@{}}stock\\ sale\\ share\\ billion\\ fall\end{tabular}           & \begin{tabular}[c]{@{}c@{}}software\\ technology\\ service\\ internet\\ launch\end{tabular}   & \begin{tabular}[c]{@{}c@{}}song\\ music\\ musical\\ wonderful\\ dance\end{tabular}                 & \begin{tabular}[c]{@{}c@{}}waste\\ awful\\ terrible\\ boring\\ poor\end{tabular}     & \begin{tabular}[c]{@{}c@{}}machine\\ learning\\ algorithm\\ optimization\\ problem\end{tabular}       & \begin{tabular}[c]{@{}c@{}}retrieval\\ document\\ query\\ search\\ base\end{tabular}       \\ \toprule
ACorEX                   & \begin{tabular}[c]{@{}c@{}}point\\ play\\ player\\ league\\ beat\end{tabular}                  & \begin{tabular}[c]{@{}c@{}}force\\ country\\ attack\\ military\\ political\end{tabular}       & \begin{tabular}[c]{@{}c@{}}billion\\ business\\ buy\\ stock\\ profit\end{tabular}              & \begin{tabular}[c]{@{}c@{}}release $(\times)$\\ technology\\ phone\\ time $(\times)$\\ space\end{tabular}                & \begin{tabular}[c]{@{}c@{}}good\\ hear $(\times)$\\ beautiful\\ music $(\times)$\\ sound $(\times)$\end{tabular}             & \begin{tabular}[c]{@{}c@{}}bad\\ money $(\times)$\\ terrible\\ poor\\ waste\end{tabular}                     & \begin{tabular}[c]{@{}c@{}}optimization\\ gradient\\ convergence\\ stochastic\\ print $(\times)$\end{tabular}               & \begin{tabular}[c]{@{}c@{}}search\\ document\\ query\\ retrieval\\ semantics\end{tabular}        \\ \midrule
AVIAD                    & \begin{tabular}[c]{@{}c@{}}robitaille $(\times)$\\ probert $(\times)$\\ howe $(\times)$\\ player\\ nhl\end{tabular} & \begin{tabular}[c]{@{}c@{}}tragedy $(\times)$\\ policy\\ serbian\\ freedom\\ unite $(\times)$\end{tabular}  & \begin{tabular}[c]{@{}c@{}}sanwa $(\times)$\\ zoete $(\times)$\\ earning\\ overprice\\ acquirer\end{tabular} & \begin{tabular}[c]{@{}c@{}}genscher $(\times)$\\ enlargement $(\times)$\\ abm $(\times)$\\ teng $(\times)$\\ chechnya $(\times)$\end{tabular} & \begin{tabular}[c]{@{}c@{}}traditional $(\times)$\\ snow $(\times)$\\ filling\\ bisque $(\times)$\\ seaweed $(\times)$\end{tabular} & \begin{tabular}[c]{@{}c@{}}email $(\times)$\\ upset\\ management $(\times)$\\ yell\\ acknowledge $(\times)$\end{tabular}   & \begin{tabular}[c]{@{}c@{}}bind\\ analytically $(\times)$\\ certify $(\times)$\\ arm $(\times)$\\ pruning\end{tabular}                    & \begin{tabular}[c]{@{}c@{}}ehr\\ healthy $(\times)$\\ progression $(\times)$\\ patient $(\times)$\\ ehrs\end{tabular} \\ \midrule
KeyETM                   & \begin{tabular}[c]{@{}c@{}}game\\ team\\ season\\ play\\ win\end{tabular}                      & \begin{tabular}[c]{@{}c@{}}people\\ government\\ person $(\times)$\\ armenian\\ law\end{tabular}     & \begin{tabular}[c]{@{}c@{}}year $(\times)$\\ percent\\ market\\ time $(\times)$\\ month $(\times)$\end{tabular}     & \begin{tabular}[c]{@{}c@{}}company $(\times)$\\ bank $(\times)$\\ japan $(\times)$\\ china $(\times)$\\ russia $(\times)$\end{tabular}        & \begin{tabular}[c]{@{}c@{}}good\\ place $(\times)$\\ great\\ time $(\times)$\\ love\end{tabular}                      & \begin{tabular}[c]{@{}c@{}}food $(\times)$\\ order $(\times)$\\ service $(\times)$\\ eat $(\times)$\\ restaurant $(\times)$\end{tabular} & \begin{tabular}[c]{@{}c@{}}function\\ estimation $(\times)$\\ distribution $(\times)$\\ parameter $(\times)$\\ efficient $(\times)$\end{tabular} & \begin{tabular}[c]{@{}c@{}}translation $(\times)$\\ user\\ search\\ annotation\\ point $(\times)$\end{tabular} \\ \midrule
ECTM                     & \begin{tabular}[c]{@{}c@{}}game\\ team\\ win\\ season\\ league\end{tabular}                    & \begin{tabular}[c]{@{}c@{}}government\\ war\\ military\\ armenian\\ attack\end{tabular}       & \begin{tabular}[c]{@{}c@{}}company\\ bank\\ percent\\ market\\ price\end{tabular}              & \begin{tabular}[c]{@{}c@{}}space\\ site\\ technology\\ station\\ network\end{tabular}                      & \begin{tabular}[c]{@{}c@{}}great\\ music $(\times)$\\ love\\ wonderful\\ amazing\end{tabular}                  & \begin{tabular}[c]{@{}c@{}}waste\\ awful\\ terrible\\ bad\\ horrible\end{tabular}                     & \begin{tabular}[c]{@{}c@{}}optimization\\ convergence\\ stochastic\\ gradient\\ function\end{tabular}                & \begin{tabular}[c]{@{}c@{}}retrieval\\ document\\ query\\ search\\ user\end{tabular}             \\
\midrule
TopicAdapt                     & \begin{tabular}[c]{@{}c@{}}game\\ team\\ win\\ season\\play\end{tabular}                    & \begin{tabular}[c]{@{}c@{}}government\\ war\\ military\\ president\\ political\end{tabular}       & \begin{tabular}[c]{@{}c@{}}percent\\ company\\ bank\\ year\\ market\end{tabular}              & \begin{tabular}[c]{@{}c@{}}company\\ technology\\ space\\ site\\ station\end{tabular}                      & \begin{tabular}[c]{@{}c@{}}excellent\\ great \\ good\\ superb\\ perfect\end{tabular}                  & \begin{tabular}[c]{@{}c@{}}waste\\ bad\\ horrible\\ crap\\ garbage\end{tabular}                     & \begin{tabular}[c]{@{}c@{}}machine\\ problem\\ algorithm\\ convergence\\ optimization\end{tabular}                & \begin{tabular}[c]{@{}c@{}}retrieval\\ document\\ search\\ query\\ semantic\end{tabular}
\\ \bottomrule
\end{tabular}
}
\vspace{-2.0mm}
\caption{Qualitative Evaluation}
\label{tab:qualitative_short}
\vspace{-5.00mm}
\end{table*}
In comparison to baselines, our method's generated topic terms are generally pertinent and simple to understand. We also note that the topics created by AcorEx have respectable interpretability (See Appendix \ref{sec:qual_eval}). However, rather than adapting to the target corpus, AcorEx's produced topics strictly converge toward the prior representation. Our approach, in contrast, tends to capture the elements of the given themes that are unique to the target corpus. AVIAD, on the other hand, has the opposite problem. It varies so widely that the subjects are incredibly challenging to comprehend. When the target corpus is balanced, the KeyETM with a similar base model (ETM) to ours performs better. For instance, the keyETM works well since the dataset 20Newsg is relatively balanced. Our model consistently outperforms the competition because it benefits from both topic-level supervision and document-level supervision from existing knowledge sources to make the topics adjusted to the target corpus while also maintaining the semantics of the given topic names. Moreover, the words from each inferred topic are more semantically related to each other than other baselines, thanks to our word-level supervision.

\subsection{Case studies}
\textbf{Case study 1}:
From Table \ref{tab:case1}, we can see our model can generate new topics from the target corpus without supervision from the source corpus. We infer the topic names by observing the top 5 words of each topic.\\
\vspace{-10pt}

\begin{table}[h]
\centering
\begin{tabular}{@{}ll@{}}
\toprule
Topic Name   & Top 5 words                         \\ \midrule
gun violence & gun, law, president, firearm, crime \\
sales        & price, sale, buy, sell, work        \\ \bottomrule
\end{tabular}
\caption{Case study 1- No supervision for target corpus-specific topic. We infer the topic names by observing the top 5 words of each topic}
\label{tab:case1}
\end{table}

\vspace{-10pt}
\noindent\textbf{Case study 2:} For this experiment, we use the AG News dataset as the source corpus and NYT corpus as the target domain. Particularly, we selected the period of attack at the Twin Towers from the New York Times corpus to investigate if the model can adapt given topics from the source corpus as well as find new topics from the target corpus.
From Table \ref{tab:case2}, we can see that the model is able to adapt relevant words for each topic related to both cases. For the topics of 911 and 9/11, we used minimal supervision by providing the topic surface names to the model. 911 is mostly related to medical emergency cases, whereas 9/11 refers to the terrorist attack. The model is able to identify the most relevant words from the target documents (NYT corpus on 9/11/2001).

\begin{table}[h]
\centering
\resizebox{1.0\linewidth}{!}{%
\begin{tabular}{@{}ll@{}}
\toprule
Topic Name & Top 5 Words                                        \\ \midrule
9/11       & terrorist, terror, terrorism, militant, terrorists \\
911        & doctor, hospital, medical, physician, nurse        \\ \bottomrule
\end{tabular}}
\caption{Case study 2- Topic name as minimal supervision for target corpus specific topic}
\label{tab:case2}
\end{table}
\vspace{-15pt}

\section{Conclusion}
\label{sec:conclusion}

In this paper, we propose a problem of adapting topics from a source corpus to a target corpus, also identifying new topics for the target corpus. Different from a recent work called coordinated topic modeling which only uses well-defined topics to describe a new corpus, we also mine new topics that represent the target corpus. For this purpose, we design a method named TopicAdapt which is based on an embedded topic model \cite{dieng2020topic} that uses three levels of supervision namely word-level supervision, topic-level supervision, and document-level supervision. An extensive experiment over a set of datasets from different domains demonstrates the superiority of the proposed model over multiple strong baselines.
\section{Limitation}
\label{sec:limitation}
The proposed model depends on two pretrained models- pretrained word-embedding for vocabulary words and pretrained language model for textual entailment during generating document-level supervision. However, for a very specific target domain, the pretarined knowledge might be appropriate. In such a case, finetuning those models on the target corpus is worth exploring for better performance. Moreover, similar to CTM \cite{akash2022coordinated},
in this paper, we assume that the reference and target corpora are from common or very similar domains. However, practically, it is very probable that we may need to transfer topic knowledge from one domain to another. Therefore, extending our model for cross-domain scenarios is also an interesting future direction.

\bibliography{anthology,custom}
\bibliographystyle{acl_natbib}

\appendix
\section{Appendix}
\label{sec:appendix}

\subsection{Related Works}
\label{sec:related}
\subsection*{Transfer Learning}
Transfer learning is a machine learning technique where a model trained on one task is used as a starting point for a model on a related but different task. This allows the model to take advantage of previously learned representations, improving the training speed and performance compared to starting from scratch. Transfer learning has become a popular technique in deep learning, where pre-trained models on large datasets can be fine-tuned for specific tasks with much smaller datasets. 

Transfer learning has been used in image classification, natural language processing (NLP), speech recognition, etc. In NLP, transfer learning refers to pre-training an NLP model on a large corpus of text, such as a general-purpose language modeling dataset, and then fine-tuning the model on a smaller, specific NLP task, such as sentiment analysis or question answering. Some examples of pre-trained models used for transfer learning in NLP include BERT \cite{devlin-etal-2019-bert}, GPT \cite{radford2018improving}, ELMo \cite{peters-etal-2018-deep}, etc.

\subsection*{Topic Modeling}
Topic modeling is a machine learning technique used to identify patterns in large collections of text data. The goal of topic modeling is to discover topics or themes that occur in a set of documents and quantify each document's relevance to these topics. There are mainly three types of topic modeling approaches: 
\begin{itemize}
    \item \textbf{Probabilistic topic modeling}: Probabilistic topic modeling uses probabilistic methods to identify topics in a collection of text data. This type of topic modeling views each document as a mixture of topics and each topic as a distribution over words. Some common probabilistic topic modeling algorithms are LDA \cite{blei2003latent}, PLSA \cite{hofmann1999probabilistic}, etc.
    \item \textbf{Neural topic modeling}: Neural topic modeling is a type of topic modeling that uses deep learning techniques to model the relationships between topics and words in a collection of text data. In neural topic modeling, the model typically consists of two components: an encoder that transforms the text data into a low-dimensional representation and a decoder that maps the low-dimensional representation back to the original space of words. The model is trained to minimize the reconstruction error between the input and the reconstructed text while also encouraging the low-dimensional representation to capture the underlying topics in the data. Some well-known neural topic models are ProdLDA \cite{miao2016neural}, NVDM \cite{srivastava2016neural}, etc.
    
    \item \textbf{User-guided topic modeling}: User-guided topic modeling is a type of topic modeling that allows the user to provide additional information or guidance to the topic modeling process. When training a model using predicted document category labels, Supervised LDA \cite{mcauliffe2007supervised} and DiscLDA \cite{lacoste2008disclda} make the assumption that each document has a label associated with it. To create topic models, several research uses word-level supervision. For instance, Dirichlet Forest \cite{andrzejewski2009latent} priors have been used to include constraints on must-link and cannot-link relationships between seed words. A seed topic distribution is used by Seeded LDA \cite{jagarlamudi2012incorporating} to learn seed-related topics under the supervision of user-supplied seed words. Finally, there is an approach called category-guided topic mining \cite{meng2020discriminative} (CatE), which considers the topics' surface names as the only supervision for mining user-interested discriminative topics.
    
\end{itemize}

\subsection{Data}
\label{sec:data_detail}
We use three datasets from news articles: 
\begin{itemize}
    \item 20 Newsgroup corpus \footnote{http://qwone.com/~jason/20Newsgroups/}: It consists of approximately 20,000 documents, each belonging to one of 20 different newsgroups.
    \item New York Times annotated corpus \cite{sandhaus2008new}:
    It contains over 1.8 million articles, each annotated with a rich set of metadata including headline, byline, date, section, and article abstract.
    \item AG’s News dataset \cite{yang2016hierarchical}:It contains over 1 million news articles, with approximately 30,000 articles per category. 
\end{itemize}

For review sentiment domain, we use:
\begin{itemize}
    \item Yelp restaurant review dataset: The dataset includes over 8 million reviews, 200,000 businesses, and 6 million users from various locations around the world.
    \item IMDB Movie Review dataset: The dataset includes 50,000 movie reviews, with 25,000 labeled as positive and 25,000 labeled as negative.
\end{itemize}

For academic articles we use: 
\begin{itemize}
    \item Arxiv Artificial Intelligence (AI) article abstracts spanning 2020-2022 \footnote{https://www.kaggle.com/Cornell-University/arxiv};
    \item Microsoft Academic Graph AI article abstracts \cite{sinha2015overview}.
\end{itemize}

\subsection{Baselines}
\label{sec:baseline_detail}
We compare our model with the following
baselines. 
\begin{itemize}
    \item GLDA: Guided LDA \cite{jagarlamudi2012incorporating} introduces bias into the generative process of LDA by utilizing topic-level priors over vocabulary based on designated seed words.
    \item Sup+LLDA: Supervised Labeled LDA is an extension of Labeled-LDA \cite{ramage2009labeled} where a label for each document is predicted from a supervised BERT  learned on annotated reference corpus.
    \item ZS+LLDA: Zero-Shot Labeled LDA is also an extension of Labeled-LDA \cite{ramage2009labeled} where a label for each document is inferred from given surface names using a Zero-Shot classification.
    \item ACorEx: Anchored CorEx uses topic correlation to learn topics with maximal information. It also uses
    user-provided seed words as anchors to bias compression of the original corpus.
    \item AVIAD: AVIAD \cite{hoang2019towards} aims to incorporate prior knowledge obtained from seed words into the model by altering the loss function to infer the desired topics.
    \item KeyETM: Keyword Assisted ETM \cite{harandizadeh2022keyword} integrates prior knowledge obtained from designated seed words.
    \item ECTM: ECTM \cite{saha2022coordinated} uses topic- and document-level supervision for topic modeling.
\end{itemize}

\subsection{Evaluation Metrics}
\label{sec:eval_metric}
\begin{itemize}
    \item Topic coherence (TC): TC is a standard measure of interpretability
    based on the average point-wise mutual
    information between randomly drawn two
    words from a document.
    \item Topic diversity (TD): TD measures the percentage of unique words in the top 25 words from all topics.
    \item Topic Quality (TQ): TQ is the product of topic coherence and topic diversity.
\end{itemize}

\subsection{Qualitative Evaluation}
\label{sec:qual_eval}
The complete qualitative result is shown in Table \ref{tab:qualitative}.

\begin{table*}[]
\centering
\resizebox{0.8\linewidth}{!}{%
\begin{tabular}{@{}c|cc|cc|cc|cc@{}}
\toprule
\multirow{2}{*}{} & \multicolumn{2}{c|}{\textbf{20Newsg}}                                                                                                                                                                    & \multicolumn{2}{c|}{\textbf{NYT}}                                                                                                                                                                                     & \multicolumn{2}{c|}{\textbf{Yelp-Senti}}                                                                                                                                                                                  & \multicolumn{2}{c}{\textbf{Arxiv-AI}}                                                                                                                                                                                            \\ 
                         & sports                                                                                         & politics                                                                                      & business                                                                                       & technology                                                                                                 & good                                                                                                    & bad                                                                                                   & ML                                                                                                                   & IR                                                                                               \\ \toprule
\begin{tabular}[c]{@{}c@{}} Reference\\Topic\\ Words\end{tabular}                    & \begin{tabular}[c]{@{}c@{}}night\\play\\sport\\player\\beat\end{tabular}                 & \begin{tabular}[c]{@{}c@{}}leader\\ election \\ attack\\ afp\\ iraqi \end{tabular}       & \begin{tabular}[c]{@{}c@{}}stock\\ sale\\ share\\ billion\\ fall\end{tabular}           & \begin{tabular}[c]{@{}c@{}}software\\ technology\\ service\\ internet\\ launch\end{tabular}   & \begin{tabular}[c]{@{}c@{}}song\\ music\\ musical\\ wonderful\\ dance\end{tabular}                 & \begin{tabular}[c]{@{}c@{}}waste\\ awful\\ terrible\\ boring\\ poor\end{tabular}     & \begin{tabular}[c]{@{}c@{}}machine\\ learning\\ algorithm\\ optimization\\ problem\end{tabular}       & \begin{tabular}[c]{@{}c@{}}retrieval\\ document\\ query\\ search\\ base\end{tabular}       \\ \toprule
GLDA                     & \begin{tabular}[c]{@{}c@{}}game\\ team\\ year $(\times)$\\ play\\ player\end{tabular}                 & \begin{tabular}[c]{@{}c@{}}people\\ time $(\times)$\\ government\\ gun\\ year $(\times)$\end{tabular}       & \begin{tabular}[c]{@{}c@{}}company\\ percent\\ year $(\times)$\\ bank\\ market\end{tabular}           & \begin{tabular}[c]{@{}c@{}}president $(\times)$\\ bush $(\times)$\\ official $(\times)$\\ united $(\times)$\\ house $(\times)$\end{tabular}   & \begin{tabular}[c]{@{}c@{}}good\\ place $(\times)$\\ food $(\times)$\\ great\\ order $(\times)$\end{tabular}                 & \begin{tabular}[c]{@{}c@{}}order $(\times)$\\ food $(\times)$\\ time $(\times)$\\ place $(\times)$\\ service $(\times)$\end{tabular}     & \begin{tabular}[c]{@{}c@{}}adversarial\\ distribution $(\times)$\\ class $(\times)$\\ function $(\times)$\\ attack $(\times)$\end{tabular}       & \begin{tabular}[c]{@{}c@{}}graph\\ search\\ user\\ class $(\times)$\\ recommendation\end{tabular}       \\ \midrule

Sup+LLDA                   & 

\begin{tabular}[c]{@{}c@{}}game\\ team\\ year $(\times)$\\ play\\ time $(\times)$\end{tabular}                 & 

\begin{tabular}[c]{@{}c@{}} people \\ government \\ kill \\ time $(\times)$ \\ year $(\times)$ \end{tabular} & 

\begin{tabular}[c]{@{}c@{}}year $(\times)$  \\ percent \\ company \\ market \\ government $(\times)$ \end{tabular}       & 

\begin{tabular}[c]{@{}c@{}}year $(\times)$ \\ time $(\times)$ \\ people $(\times)$ \\ president $(\times)$ \\ official $(\times)$\end{tabular}      & 

\begin{tabular}[c]{@{}c@{}} place $(\times)$ \\ food $(\times)$ \\ good \\ great \\ service $(\times)$ \end{tabular} &

\begin{tabular}[c]{@{}c@{}} food $(\times)$ \\ order $(\times)$ \\ place $(\times)$ \\ service $(\times)$ \\ time $(\times)$ \end{tabular}               & 

\begin{tabular}[c]{@{}c@{}}demonstrate $(\times)$ \\ problem $(\times)$ \\ feature \\ training \\ neural \end{tabular}          & 

\begin{tabular}[c]{@{}c@{}} retrieval \\ exist $(\times)$ \\ demonstrate $(\times)$ \\ representation \\ feature $(\times)$ \end{tabular}

\\ \midrule
ZS+LLDA                   & \begin{tabular}[c]{@{}c@{}}game\\ team\\ year $(\times)$\\ play\\ player\end{tabular}                 & \begin{tabular}[c]{@{}c@{}}people\\ time $(\times)$\\ government\\ year $(\times)$\\ point $(\times)$\end{tabular} & \begin{tabular}[c]{@{}c@{}}year $(\times)$\\ percent\\ company\\ market\\ lead $(\times)$\end{tabular}       & \begin{tabular}[c]{@{}c@{}}year $(\times)$\\ time $(\times)$\\ american $(\times)$\\ official $(\times)$\\ today $(\times)$\end{tabular}      & \begin{tabular}[c]{@{}c@{}}food $(\times)$\\ place $(\times)$\\ good\\ great\\ service $(\times)$\end{tabular}               & \begin{tabular}[c]{@{}c@{}}food $(\times)$\\ order $(\times)$\\ place $(\times)$\\ service $(\times)$\\ time $(\times)$\end{tabular}     & \begin{tabular}[c]{@{}c@{}}efficient $(\times)$\\ reduce $(\times)$\\ number $(\times)$\\ leverage $(\times)$\\ module $(\times)$\end{tabular}          & \begin{tabular}[c]{@{}c@{}}retrieval\\ search\\ user\\ document\\ query\end{tabular}             \\ \midrule
ACorEX                   & \begin{tabular}[c]{@{}c@{}}point\\ play\\ player\\ league\\ beat\end{tabular}                  & \begin{tabular}[c]{@{}c@{}}force\\ country\\ attack\\ military\\ political\end{tabular}       & \begin{tabular}[c]{@{}c@{}}billion\\ business\\ buy\\ stock\\ profit\end{tabular}              & \begin{tabular}[c]{@{}c@{}}release $(\times)$\\ technology\\ phone\\ time $(\times)$\\ space\end{tabular}                & \begin{tabular}[c]{@{}c@{}}good\\ hear $(\times)$\\ beautiful\\ music $(\times)$\\ sound $(\times)$\end{tabular}             & \begin{tabular}[c]{@{}c@{}}bad\\ money $(\times)$\\ terrible\\ poor\\ waste\end{tabular}                     & \begin{tabular}[c]{@{}c@{}}optimization\\ gradient\\ convergence\\ stochastic\\ print $(\times)$\end{tabular}               & \begin{tabular}[c]{@{}c@{}}search\\ document\\ query\\ retrieval\\ semantics\end{tabular}        \\ \midrule
AVIAD                    & \begin{tabular}[c]{@{}c@{}}robitaille $(\times)$\\ probert $(\times)$\\ howe $(\times)$\\ player\\ nhl\end{tabular} & \begin{tabular}[c]{@{}c@{}}tragedy $(\times)$\\ policy\\ serbian\\ freedom\\ unite $(\times)$\end{tabular}  & \begin{tabular}[c]{@{}c@{}}sanwa $(\times)$\\ zoete $(\times)$\\ earning\\ overprice\\ acquirer\end{tabular} & \begin{tabular}[c]{@{}c@{}}genscher $(\times)$\\ enlargement $(\times)$\\ abm $(\times)$\\ teng $(\times)$\\ chechnya $(\times)$\end{tabular} & \begin{tabular}[c]{@{}c@{}}traditional $(\times)$\\ snow $(\times)$\\ filling\\ bisque $(\times)$\\ seaweed $(\times)$\end{tabular} & \begin{tabular}[c]{@{}c@{}}email $(\times)$\\ upset\\ management $(\times)$\\ yell\\ acknowledge $(\times)$\end{tabular}   & \begin{tabular}[c]{@{}c@{}}bind\\ analytically $(\times)$\\ certify $(\times)$\\ arm $(\times)$\\ pruning\end{tabular}                    & \begin{tabular}[c]{@{}c@{}}ehr\\ healthy $(\times)$\\ progression $(\times)$\\ patient $(\times)$\\ ehrs\end{tabular} \\ \midrule
KeyETM                   & \begin{tabular}[c]{@{}c@{}}game\\ team\\ season\\ play\\ win\end{tabular}                      & \begin{tabular}[c]{@{}c@{}}people\\ government\\ person $(\times)$\\ armenian\\ law\end{tabular}     & \begin{tabular}[c]{@{}c@{}}year $(\times)$\\ percent\\ market\\ time $(\times)$\\ month $(\times)$\end{tabular}     & \begin{tabular}[c]{@{}c@{}}company $(\times)$\\ bank $(\times)$\\ japan $(\times)$\\ china $(\times)$\\ russia $(\times)$\end{tabular}        & \begin{tabular}[c]{@{}c@{}}good\\ place $(\times)$\\ great\\ time $(\times)$\\ love\end{tabular}                      & \begin{tabular}[c]{@{}c@{}}food $(\times)$\\ order $(\times)$\\ service $(\times)$\\ eat $(\times)$\\ restaurant $(\times)$\end{tabular} & \begin{tabular}[c]{@{}c@{}}function\\ estimation $(\times)$\\ distribution $(\times)$\\ parameter $(\times)$\\ efficient $(\times)$\end{tabular} & \begin{tabular}[c]{@{}c@{}}translation $(\times)$\\ user\\ search\\ annotation\\ point $(\times)$\end{tabular} \\ \midrule
ECTM                     & \begin{tabular}[c]{@{}c@{}}game\\ team\\ win\\ season\\ league\end{tabular}                    & \begin{tabular}[c]{@{}c@{}}government\\ war\\ military\\ armenian\\ attack\end{tabular}       & \begin{tabular}[c]{@{}c@{}}company\\ bank\\ percent\\ market\\ price\end{tabular}              & \begin{tabular}[c]{@{}c@{}}space\\ site\\ technology\\ station\\ network\end{tabular}                      & \begin{tabular}[c]{@{}c@{}}great\\ music $(\times)$\\ love\\ wonderful\\ amazing\end{tabular}                  & \begin{tabular}[c]{@{}c@{}}waste\\ awful\\ terrible\\ bad\\ horrible\end{tabular}                     & \begin{tabular}[c]{@{}c@{}}optimization\\ convergence\\ stochastic\\ gradient\\ function\end{tabular}                & \begin{tabular}[c]{@{}c@{}}retrieval\\ document\\ query\\ search\\ user\end{tabular}             \\
\midrule
TopicAdapt                     & \begin{tabular}[c]{@{}c@{}}game\\ team\\ win\\ season\\play\end{tabular}                    & \begin{tabular}[c]{@{}c@{}}government\\ war\\ military\\ president\\ political\end{tabular}       & \begin{tabular}[c]{@{}c@{}}percent\\ company\\ bank\\ year\\ market\end{tabular}              & \begin{tabular}[c]{@{}c@{}}company\\ technology\\ space\\ site\\ station\end{tabular}                      & \begin{tabular}[c]{@{}c@{}}excellent\\ great \\ good\\ superb\\ perfect\end{tabular}                  & \begin{tabular}[c]{@{}c@{}}waste\\ bad\\ horrible\\ crap\\ garbage\end{tabular}                     & \begin{tabular}[c]{@{}c@{}}machine\\ problem\\ algorithm\\ convergence\\ optimization\end{tabular}                & \begin{tabular}[c]{@{}c@{}}retrieval\\ document\\ search\\ query\\ semantic\end{tabular}
\\ \bottomrule
\end{tabular}
}
\vspace{-2.0mm}
\caption{Qualitative Evaluation}
\label{tab:qualitative}
\vspace{-5.00mm}
\end{table*}

\end{document}